# Correspondent Banking Networks: Theory and Experiment

**Nima Safaei**

*Data Science and Analytics, Customer Insights, Data & Analytics, Scotiabank, Toronto; ON; Canada,*
*nima.safaei@scotiabank.com,*

**Ivan A. Sergienko**

*Riskfuel Analytics Inc., Toronto, ON, Canada, ivan@riskfuel.com*

ABSTRACT

We employ the mathematical programming approach in conjunction with the graph theory to study the structure of correspondent banking networks. Optimizing the network requires decisions to be made to onboard, terminate or restrict the bank relationships to optimize the size and overall risk of the network. This study provides theoretical foundation to detect the components, the removal of which does not affect some key properties of the network such as connectivity and diameter. We find that the correspondent banking networks have a feature we call *k*-accessibility, which helps to drastically reduce the computational burden required for finding the abovementioned components. We prove a number of fundamental theorems related to *k*-accessible directed graphs, which should be also applicable beyond the particular problem of financial networks. The theoretical findings are verified through the data from a large international bank.

## 1. INTRODUCTION

Correspondent banking (CB) network refers to a network of financial institutions providing cross-border payment services for customers through different channels such as SWIFT, Fedwire, etc. Connections in the CB network represent existing relationships among banks and customers. Through the CB network, banks and their customers can access financial services in different jurisdictions and provide cross-border payment services to their customers, supporting, among other things, international trade and financial inclusion. In addition, most payment solutions that do not involve a bank account at the customer level (*e.g.* remittances) rely on CB for the actual transfer of funds [1].

### 1.1. The Aspirations of the Work

In today's environment, financial institutions globally are at risk of being used by criminal organizations to launder money and by terrorist groups to facilitate the financing of their activities. Consequently, banks globally are facing increasing scrutiny from both regulators and clients. Until recently, banks have maintained a broad network of correspondent relationships, but this situation is changing due to the global financial crisis of the last decade and economic slowdown. In particular, the profitability of the CB network is under pressure as banks need to comply with more regulation concerning risks such as money-laundering, terrorist financing, credit loss, reputation, *etc*. For example, legislative sanctions acts may prohibit the banks from creating, maintaining, or holding a correspondent bank account with a foreign bank when the foreign bank holds or maintains a correspondent bank account with a bank or within a country considered illicit under the act. Increasing inter-bank competition due to low growth conditions in mature markets is another issue facing banks.

A report published by the Boston Consulting Group in 2011 revealed that global payments volumes are forecasted to grow at an average compound rate of 9% per year through to 2020 [2]. They estimated that the global payments market will be worth $782 trillion in noncash transaction value and $492 billion in transaction revenues by 2020. Accounting for 2.5% of the global cash volume in 2010, cross-border payments generated 10% of the revenue, and by 2020 are forecasted to account for 3.5% of the volume and 16% of revenue. So in 2011, the Celent consulting firm reported that the cost of compliance with Anti-Money Laundering (AML) regulations is increasing at 7.9% per year [3]. This report resulted from an exclusive survey of AML compliance departments at more than 75 financial institutions globally. These factors force the banks to optimize their CB networks by potentially adjusting the number of relationships they maintain. When establishing new relationships, the banks must consider various profits, risks and cost criteria. [4].

This study is the first to examine the structure of the CB network from a graph theory perspective. This should not be confused with other banking network problems in economics and finance such as *Systematic Risk Assessment* of financial networks, a well-understood topic in the literature [5, 6]. *Systematic Risk Assessment* refers to analyzing the risk of interbank liabilities and assets in the form of unanticipated changes in macroeconomic variables, market-wide illiquidity or chain reaction defaults [7]. Other example to quantify the systematic risk is the networks of volatility spillovers, for example, for modeling the stock market relationships [8]. In a similar manner, *Systemic* Risk Assessment refers to assessment of the transmission of risk through a network of interbank obligations, in which the shock typically occurs only in one area of the network, and then spreads out to other areas through financial contagion (inability of a bank to make payments reduces the liquidity of its direct counterparties, triggering illiquidity and insolvency). In above-mentioned literature, the network science has received growing attention to understand the behavior and dynamics of financial markets, and describing the interconnectedness between financial agents, especially, after the recent global financial crisis. However, the literature around CB networks is limited to a few reports and patents published since 2012 (see Section 1.2), revealing that the CB network remains little discussed and understood by academics and institutional organizations. Consequently, there is little theoretical foundation and guidance for practitioners and academicians alike [9]. One possible reason for this is rooted in the sheer complexity of *Directed Graphs (digraphs)*. The CB networks must be modeled as digraphs; however, there has been much less success in the study of the spectra of directed graphs than of undirected graphs, perhaps because the associated non-symmetric adjacency matrices are not necessarily diagonalizable. Moreover, real-world networks rarely have fine characteristics of directed graphs (such as *k*-connectedness, *k*-regularity, *k*-order, eulerian-ness, tournament, Pancyclicity,...) to be analyzed through the classical methods of graph theory.

### 1.2. Literature Review

We found only three peer-reviewed studies particularly focused on CB networks. Koko (2012) proposed a method for determining the *minimum degree of separation* between two banks [10]. As a result of regulation, it is important for a bank to investigate foreign institutions to determine the degree of separation the bank has from an illicit entity via a foreign correspondent bank. From a graph theory perspective, the degree of separation of any two banks may be computed as the minimum number of edges (relationships) in any path linking the two banks. The proposed method considers the CB network with directional relationships so that the associated bank may have correspondent bank accounts with other banks (Nostro relationship) or vice versa (Vostro relationship). Using this method, one can study potential scenarios of violating the regulation by a bank in the CB network. This study used the concept of *Bridge* banks, which are a small set of banks that facilitate the connection between two distinct groups of banks. Hampapur et al. (2015) provided a design of an optimization engine that identified the best routes for a set of transactions to minimize cost, maximize revenue and reduce risk [11]. In a similar context, Soramaki et al. (2007) investigated the topology of interbank payment flows. They analyzed a stochastic network created by daily transaction data from the Fedwire® Funds Service. They found that the network is degree distribution scale-free and compact despite low connectivity. Moreover, they revealed that the properties of the network changed considerably in the immediate aftermath of the attacks of September 11, 2001 and the North American blackout of August 14, 2003 [12].

### 1.3 Contribution

The above research works studied only a very specific aspect of the CB network with certain prior knowledge, e.g., the illicit entity or bridge set is known in advance. However, to optimize the CB network in a general sense, we have to first learn the key properties hidden in the network structure from a connectivity point of view. These properties enable us to understand the network topology with the aim of reducing the computational burden of network size optimization, mining network for specific components, analyzing what-if scenarios for off-boarding/on-boarding the banks, and reducing the risk of not meeting regulation. To this end, we consider a deterministic CB network associated to a longer time window (e.g., year); instead of daily stochastic networks as considered in Soramaki et al. (2007). We study the network only from a topological (connectivity) point of view; regardless of the amount of money and volume of transactions between pair entities. That is, we suppose an *unweighted* digraph in which all connections (edges) and entities (nodes) have same weight. This assumption enables us to study the structural properties of CB network where the importance of each entity only depends on its neighbor subnetwork. These properties help us, for example, to find the components the removal of which destroys the network connectivity. This is the case when off-boarding of elicit entities is of interest. As another example, the extracted properties are also enabling us to optimize the network where the only available options are to *keep*, *remove*, or *restrict* the entities. For a better clarification, we split the notion of network optimization in two: 1- *network size optimization*: removing some intermediary banks from the CB network without direct impact to banks' end customers (individuals, businesses or government entities) sending payments to each other. That is, after review, of each sender-receiver pair there is still a route through the network connecting them (re-routing some transactions). Potentially reducing banks in the network allows for indirect customer impact, *e.g.* increased fees due to more banks involved in sending a particular payment. 2- *Interconnectivity optimization:* allowing senders or receivers not to participate in certain activities in the network. This is equivalent to removing some connections (instead of entities) which is interpreted as restriction on relationship with some banks. This would inevitably lead to the loss of CB revenue, which must be balanced with the associated reduction in risk. While the former scenario is the key motivation of this study, investigating the latter scenario is beyond the scope of this study and recommended as future work. In this paper, we focus exclusively on evaluating the impact of removing the components (a set of entities) on some key properties of the network such as *connectivity* and *diameter*. Although the mathematical apparatus we develop here will also be useful to address the more complicated problem of Interconnectivity optimization or weighted network in which not every relationship or entity has the same risk/cost of removal.

## 1.4 Methodology

Nowadays, large-scale graph data is being generated in a variety of real-world applications, from social networks to co-authorship networks, and from financial networks to road traffic networks. Thus, graph mining methods are receiving prominent attention to study high-order network structures with the aim of prediction or anomaly detection. In this study, our focus is on digraphs that appear in many application scenarios including, but not limited to, online social networks such as Twitter (e.g., referring to follower relations or to actual communication among users) and financial networks. In fact, recent studies reveal that most real-world networks are directed [13]. Our findings are also applicable to many directed graphs beyond financial networks.

The goal of the majority of graph mining methods is to find structure-rich subgraphs, *i.e.* components containing most of the information about the entire graph. A key challenge associated with finding such structures is a prohibitive computational cost. The problem belongs to the family of local graph clustering problems, for which efficiently identifying a low-conductance cut without exploring the entire graph is of interest. One well-known structure-rich component is the *strongly connected component (SCC)* which comprises all entities in the graph that can reach each other through a directed path. Empirical studies reveal that the majority of real-world directed networks contain a Giant SCC [12, 14, 15]. For example, Soramaki et al (2007) showed that 0.3% of entities in the Fedwire® network form a densely connected sub-graph (clique) to which the remaining nodes connect. Thus, finding such structure-rich components could be a decent starting point to study the structural properties of the CB network.

To this end, we model the CB network as a directed graph (*digraph*) using the historical data on SWIFT (Society for Worldwide Interbank Financial Telecommunications) cross-border payments. Our network analysis indicates the existence of a small set of banks in the CB network, each of which has access to, on average, $k$ other banks with a negligible variance. We also find that this set contains a unique SCC so that its size and density are tied up with the above-mentioned variance. We refer to this prominent phenomenon as "$k$-accessibility". Our analysis also reveals the lack of correlation between the accessibility of a vertex and its out-degree in the CB network, meaning that the vertices with a higher out-degree do not necessarily have access to a larger segment of the network. Moreover, the CB network is essentially sparse and the $k$-accessibility feature is not due to a dense network with a large number of connections. The key explanation for this feature is child-parent relationships between the banks. That is, for the most part, the branches are transferring money through the headquarters or main regional offices. However, this behaviour is not always the case as a bank's foreign branches are also allowed to independently do business with other institutions. The child-parent relationships also result in too many 2-cycle components and consequently in a non-oriented digraph it becomes more difficult to study the structural properties of the network. We will argue that the network optimization problem can be reduced to finding a maximal subset of entities that belong to the SCC such that the removal of these entities still preserves some important properties of SCC such as strong connectivity and diameter.

**1.5 Assumptions and Structure of the Paper**

Our study is supported by the following assumptions inspired by common practice, business and regulatory rules:

1. To create the CB network, we used a subset of SWIFT payment data, as viewed by Scotiabank, a large international bank headquartered in Toronto, Canada, with operations in nearly 50 countries. The bank can see all the transactions in which it is either a correspondent or intermediary bank. Each row of data represents a specific payment amount for a specific sender- receiver pair.

2. Due to a high variability of network traffic, the CB network undergoes assessment on a periodic basis. We considered a realistic scenario where only the payments over the last year are examined. The correspondent relationships and other factors affecting the network structure (e.g., banks merging) are assumed to remain unchanged.

3. The revenue obtained from the network is a direct function of the number and amount of payments. However, the temporal pattern of payments does not matter. Thus, we consider this to be a deterministic network.

4. The transactions between the banks are made through Vostro accounts in the currency of the country where the money is on deposit. Thus, any existing relationship represents an existing Vostro account with a specific currency. It is assumed that the cost of closing a Vostro account is negligible. Likewise, the cost of opening a new Vostro account with a bank that already has relationships through other currencies will be negligible too. This is because there is already a relationship agreement between two institutions. Opening a new Vostro account is for finding alternative routes for transactions which may be affected by adjusting flows through the network.

5. The network analysis is performed at the bank branch (child) level. Branches of a parent bank are geographically distributed across the world. Each branch may work with other banks independently without being monitored by the parent. One possible reason is that the Vostro accounts do not necessarily have centralized ownership.

6. The routing of transactions in the CB network follows a bifurcated pattern, meaning that the transaction does not follow a unique path between each sender-receiver pair. The routing of transactions is not fixed and may be changed depending on various factors.

The paper is structured as follows. We begin in Section 2 with graphical representation of the CB network and a brief review of fundamental definitions of digraphs. In Section 3, the structural properties of the CB network will be studied through a number of proved theorems and inferences. In Section 4, we provide evidence to verify the ideas and interpret the theorems proposed in Section 3 through real-world datasets. Finally, we conclude in Section 5.

## 2. NETWORK DESIGN AND DEFINITIONS

We represent the CB network as *digraph* $G(V, E)$ where the vertex set $V = \Psi \cup \Lambda$ is union of *customer set* $\Psi$ and *bank set* $\Lambda$. The edge set $E = E_1 \cup E_2$ is union of *correspondent* relationships $(E_1: \Psi \leftrightarrow \Lambda)$ and *interbank* relationships $(E_2: \Lambda \to \Lambda)$. More specifically, two vertices are connected with a directed edge if there was either at least one *payment* between a customer (sender/receiver) and a correspondent bank or at least one *transaction* between two banks. Thus, we turn the historical transactions into the existing relationships through graph $G$. $l^{th}$ observation represents a payment for pair *sender-receiver* $(s_l, r_l)$; $s_l, r_l \in \Psi$ to be transferred through path $P_l \subset \Lambda$ where $P_l$ is an ordered series of banks.

A digraph is *acyclic* if and only if it has no *strongly connected* subgraphs with more than one vertex. A subgraph is said to be strongly connected if every vertex is reachable from every other vertex. In our study, $G$ is assumed to be acyclic at the customer level but not at the bank level. However, the *condensation* of $G$ (each SCC is contracted to a single vertex) is acyclic, meaning that a customer can be either sender or receiver but not both at the same time. If this is the case, we add dummy vertices. Thus, $G$ is represented as a *weakly connected source-sink graph* with one-way flow from source nodes (senders) to sink nodes (receivers). We say "*weakly connected*" since even though the graph is connected there is not always a path from each sender to each receiver. The following definitions are provided to facilitate our discussion on network analytics over the next section. Some fundamental definitions are revisited from the graph theory.

***Adjacency Matrix***: the adjacency matrix of $G$, $A_G$, is a binary and non-symmetric square matrix where $A_G(i,j) = 1$ means vertex *i* is adjacent to *j* or *j* is directly accessible from *i*.

***Vertex Degree***: We define $I_v$ as *In-degree* of vertex $v$ (the number of inward edges to vertex $v$) and $O_v$ as *Out-degree* of vertex $v$ (the number of outward edges from vertex $v$) given by

$$I_v = I_v^\Psi + I_v^\Lambda = \sum_{c \in \Psi} A_G(c, v) + \sum_{b \in \Lambda} A_G(b, v),$$

$$O_v = O_v^\Psi + O_v^\Lambda = \sum_{c \in \Psi} A_G(v, c) + \sum_{b \in \Lambda} A_G(v, b).$$

*Vertex Role*: according to the above definitions, a *sender* is treated as a node with zero in-degree and a *receiver* as a node with zero out-degree. Bank $b \in \Lambda$ is a *correspondent sender* if $|I_v^\Psi| > 0$, a *correspondent receiver* if $|O_v^\Psi| > 0$, or a correspondent bank in general if $|I_v^\Psi| \times |O_v^\Psi| > 0$. Bank $b \in \Lambda$ is an *intermediary* bank if $I_v^\Lambda \times O_v^\Lambda > 0$. By these definitions, a bank may have a combination of the above roles. According to Assumption 4, the correspondent banks are not desired for evaluation since they are directly connected to the end customers. Therefore, we restrict these activities to all non-correspondent intermediary banks $v$ where $|I_v^\Psi| = |O_v^\Psi| = 0$. The *bridge set* is defined as a minimal set of banks, the removal of which divides the network into two disjointed sub-networks of senders and receivers. Figure 1 displays a schematic representation of the CB network as a source-sink network. As depicted in this figure1, the bridge set may not be unique.

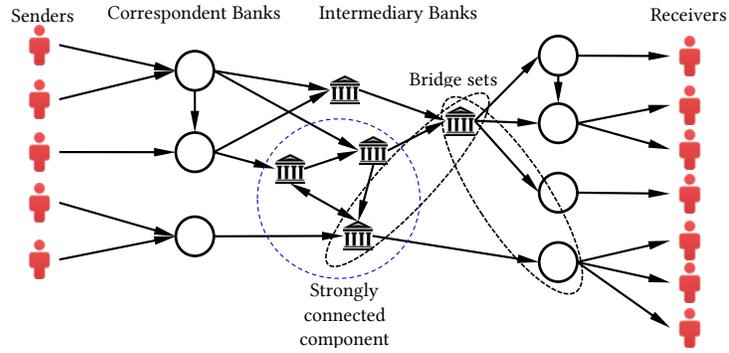

**Figure 1**. CB Network as a source-sink graph representation

*Vertex Eccentricity:* is defined as the maximum distance (length of the shortest path) from a given vertex to any other vertex in the digraph.

*Digraph Diameter:* is defined as the maximum distance between any pair of vertices or equivalently the maximal vertex eccentricity in the digraph.

*K-Clan Digraph*: is an SCC with diameter *k* (any vertex is reachable through a shortest path of size at most *k* from any other vertex). This definition is analogous with Milgram's experiment on 6-degrees *of separation (or small-world) theory* in social networks [16].

*Vertex Accessibility*: Let $\Theta_v$ be the set of all *successor* vertices accessible (directly or indirectly) through vertex $v$. This set can be obtained by the following forward breadth-first search (BFS) procedure as a traversal method.

```
Θ_v = ∅; S = {j ∈ Λ | (v,j) ∈ E}; ecc = 0
WHILE(S ≠ ∅) {
    Θ_v = Θ_v ∪ S
```

$$S = \{j \in \Lambda \mid (i,j) \in E;\ i \in \Theta_v;\ j \neq \Theta_v\}$$
$$ecc = ecc + 1$$
}

We define the size of set $\Theta_v$ as *Accessibility* of vertex $v$ where $acc(v) = |\Theta_v|$. In the above procedure, $ecc$ calculates the *eccentricity* of vertex $v$; $ecc(v)$, where $\max_v ecc(v)$ represents the digraph dimeter. If we make sure that the digraph is an SCC, then it is also $k$-Clan where $k = \max_v ecc(v)$.

*Strongly Connected Component (SCC):* maximal subgraph (maximal subset of the vertices) in a directed graph such that for every pair of vertices $u, v$ in the subgraph, there is a directed path from $u$ to $v$ and a directed path from $v$ to $u$.

*Strongly Connected Spanning Subgraph (SCSS)*: smallest subset of the edges in a given digraph preserving the strong connectivity [17]. It is also referred as the *Minimal Strongly Connected Component* [18].

*Circumference*: the length of longest cycle in the digraph [19].

***k*-accessible Digraph**: Digraph $D$ is $k$-accessible, if it contains *core* $G$; $G \subseteq D$, in which each vertex has a path to, on average, $k$ other vertices in $D$ where the variance of vertex accessibility in core $G$ is negligible. In other word, a digraph is $k$-accessible if $acc(v) \approx k\ \forall v \in G$ so that $\mu_{acc}^G = k$ and $\sigma_{acc}^G \approx 0$ where $\mu_{acc}^G$ and $\sigma_{acc}^G$ represents the mean and variance of vertex accessibility in $G$. A $k$-accessible digraph represents the complete digraph if $k = N - 1$ and $G = D$. In cases in which $k$ is not unique, the largest $k$ will be considered. For example, if $D$ is a complete digraph, it is $k$-accessible for $k = 1, 2, \ldots, N - 1$.

This feature is also analogous with the *k-Plex* definition – a maximal subgraph in which each vertex is connected to at least $N - k$ other vertices, where $N$ is the order of induced subgraph. However, $k$-Plex considers the direct but not indirect accessibility. Figure 2 represents an example of 5-accessible CB network. In the next section, we investigate the characteristics of $k$-accessible digraphs through some fundamental theorems. We will show that each $k$-accessible has a SCC where $SCC \subseteq G$ so that the size of SCC is correlated to $\sigma_{acc}$.

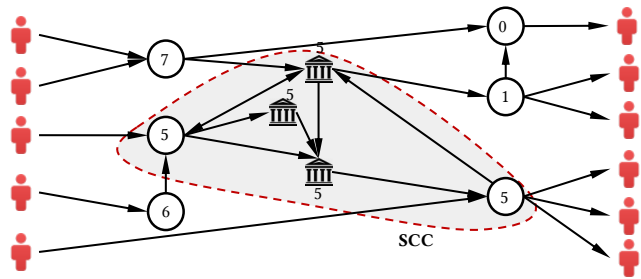

**Figure 2.** 5-accessible digraph – the numbers inside or alongside the vertex represent $acc(v)$

*Minimal K-Clan* (M$^k$C) *Digraph:* an $k$-*Clan* digraph with minimal number of edges so that by removing any edge, the diameter exceeds $k$.

*Minimal $k$-accessible* (M$^k$A) *Digraph*: an $k$-*accessible* digraph with minimal number of edges.

## 3. NETWORK ANALYTICS

Sampling from the real data reveals that the $k$-accessibility is a dominant feature of CB networks. Thus, the aim of this section is first to discover the structural properties of $k$-*accessible* digraphs through a number of theorems and then show how these theoretical inferences can be employed for the network optimization. Since finding different types of structure-rich components (e.g., SCC) in digraphs has been proved to be NP-Complete [20]; the couple of theorems are proposed to mitigate the complexity of finding such components in $k$-*accessible* digraphs.

The first two theorems emphasize that any $k$-*accessible* digraph must be cyclic having at least one SCC. Through the third theorem, we prove the uniqueness of SCC and measure its magnitude within the entire CB network. We argue that the search space for network size optimization should be limited to the SCC since all entities out of SCC connect the customers to the rest of the network and therefore should not be touched. However, preserve connectivity through the entire network, the strong connectivity of SCC should be preserved during the optimization phase.

The last three theorems help understand under which conditions the SCC could be a *k*-Clan. It is a common practice in cross-border payments to try to reduce the number of transitional banks between a sender and a receiver in order to keep the fees, charges and transaction turnaround time minimized. This is equivalent to setting an upper bound on the *diameter* of the SCC. According to the historical payments, this upper bound is equal 6; meaning that having a 6-Clan SCC should be a constraint in network optimization problem. The classical method for establishing whether a SCC is a $k$-Clan is measuring the distance between all vertex pairs or equivalently finding $\max_v ecc(v)$ in BFS procedure. Given a digraph $D(V,E)$, the complexity of BFS procedure is $O(|V|^2)$ for dense graphs or $O(|V| + |E|)$ for spars graphs. Even though we use the BFS procedure once during pre-processing task to calculate the vertex accessibility, it may not be computationally tractable to apply on large-sized SCC for verifying the $k$-Clan feature. Instead, the last three theorems enable us to verify $k$-Clan feature only through the minimal information from the graph connectivity matrix (i.e., vertex out-degree) without the need for deep exploration of the whole network.

While the first two theorems are focused on *connectivity*, the last three are focused on diameter of SCC. Having these, we will discuss that the best optimization practice to reduce the network size in an $k$-*accessible* network is to find a set of entities belong SCC, the removal of which still preserves the strong connectivity and diameter of SCC. From now on, we ignore the end customers as individual entities and suppose that they are only the sources and sinks of flow in the network. Thus, we use "digraph" as a short term for *bank subgraph* of CB network; a connected loopless digraph without multiple edges and with an order greater than three. Without loss of generality, the concepts of $M^kC$ and $M^kA$ digraphs will be used as basis to prove the theorems. It is obvious that any statement proved for minimal digraphs can be generalized for actual ones.

### 3.1. Inferences on $M^kA$ Digraphs

**Theorem 1:** In an acyclic digraph of order $N$, the average of vertex accessibility does not exceed $\frac{N-1}{2}$.

**Proof:** Consider digraph $D(V, E)$ with $|V| = N$ vertices labeled in ascending order from left to right; as shown in Figure 3. Since the digraph is acyclic, all edges are directed in ascending order of vertex labels.

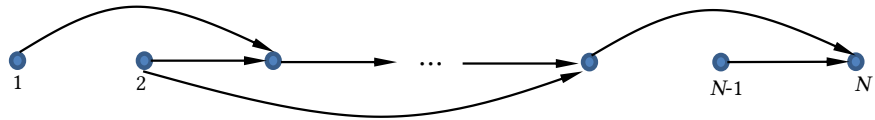

**Figure 3.** Line representation of acyclic digraph

Given arbitrary vertices $i, j \in V$, if vertex $j$ is accessible from $i$, then no vertex $v$ accessible from $j$ is adjacent to $i$. As a generalization, all vertices accessible from $j$ are not adjacent to any of the predecessors of $j$. Thus, $acc(i)$ and $acc(j)$ cannot be equal because all vertices accessible from $j$ are also accessible from $i$. Following this, the average of vertex accessibility will be maximized if each vertex $i$ has access to all its successors $i+1, \ldots, N$. Meaning that, $acc(i) = N - i$ and consequently the maximal $\mu_{acc}^D$ is obtained as $\frac{1}{N}\sum_{i=1}^{N-1}(N-i) = \frac{N-1}{2}$.

**Corollary 1:** Digraph $D$ with $\mu_{acc}^D > \frac{N-1}{2}$ cannot be acyclic.

**Theorem 2:** Each $M^kA$ digraph $D$ with $(k \geq \frac{N}{2})$ has minimum $\sigma_{acc}^D$ and at least one SCC of size $k+1$.

**Proof:** First we suppose $D$ is acyclic without SCC and try to construct an $M^kA$ digraph of order $N$ so that $\sigma_{acc}^D$ is minimized. Then, we show that turning the acyclic $M^kA$ digraph to a cycle form results in further reduction in $\sigma_{acc}^D$. We assume that $k \leq \frac{N-1}{2}$ because, according to Corollary 1, the digraph is already cyclic for $k > \frac{N-1}{2}$. Again, we assume a structure similar to Figure 3 in which all edges are directed in ascending order of vertex labels. Let $x_e$ be the number of vertices with accessibility equals $e$ where $0 \leq e \leq N - 1$. Since the digraph is assumed to be acyclic, there exists at least one sink (a vertex without outward edge) and thus $1 \leq x_0 \leq N - 1$. Accordingly, $x_1$ will be bounded to $N - x_0$. By induction, the upper bound $N - \sum_{l=0}^{e-1} x_l$ will be imposed on $x_e \ \forall e > 0$. We therefore propose the following linear integer programming model to construct an acyclic $M^k A$ with the least variance of vertex accessibility.

$$\min \sigma_{acc}^D = \frac{1}{N} \sum_{e=0}^{N-1} x_e (k-e)^2, \tag{1}$$

Subject to:

$$1 \leq x_0 \leq N - 1, \tag{1.1}$$

$$x_e \leq N - \sum_{l=0}^{e-1} x_l; \quad 1 \leq e \leq N - 1, \tag{1.2}$$

$$\sum_{e=0}^{N-1} x_e = N, \tag{1.3}$$

$$k - \frac{1}{2} < \frac{\sum_{e=1}^{N-1} e x_e}{N} < k + \frac{1}{2}, \tag{1.4}$$

$$x_e \leq M^+ \times x_{e-1} \ \forall e > 0. \tag{1.5}$$

$$x_e \in \mathbb{Z}^+.$$

Objective function (1) calculates the variance of vertex accessibility where $k$ is given. Constraints (1.1) and (1.2) impose upper bound on decision variables. Equality (1.3) guarantees the order of the digraph. Inequality (1.4) imposes that the average accessibility of vertices must be enough close to $k$ so that $\left[\frac{\sum_{e=0}^{N-1} e x_e}{N}\right] = k$ is assured. Finally, Inequality (1.5) guarantees the digraph connectivity where $M^+$ is a large positive number. Model (1) can be optimally solved using the classical MIP methods. Through benchmarking on various values of $N$ and $k$, we realize that Model (1) has the following optimality structure:

$$x_0 = x_1 = \cdots x_k = 1, \ \sum_{e=k+1}^{N-1} x_e = N - k - 1,$$

with the most frequent solution

$$X^* = (x_0 = 1, \ldots, x_k = 1, x_{k+1} = N - k - 1, x_{k+2} = 0, \ldots, x_{N-1} = 0).$$

Solution $X^*$ in fact represents a spanning tree as depicted in Figure 4.1. The optimal variance of vertex accessibility associated to this the optimal solution is obtained as

$$\sigma_{acc}^D = \frac{1}{N} \left( \sum_{e=0}^{k} (k-e)^2 + (N-k-1)(k-(k+1))^2 \right) = \frac{(k+1)(2k+1)k}{6N} - \frac{k+1}{N} + 1, \tag{2}$$

and the average of vertex accessibility is obtained as $\sigma_{acc}^D = (k+1)\left(1 - \frac{k+2}{2N}\right)$ where (1.4) imposes that $k - \frac{1}{2} < (k+1)\left(1 - \frac{k+2}{2N}\right) < k + \frac{1}{2}$ and thereby

$$N < (k+1)(k+2) < 3N, \tag{3}$$

is a necessary condition to guarantee (1.4).

Now we show that turning an acyclic digraph to a cyclic one will reduce $\sigma_{acc}^D$ below the optimal solution of Model (1) so that the average will remain slightly unchanged. To this end, we turn the optimal solution constructed by model (1) to a cyclic digraph by creating a cycle component. So, we add an edge from the sink to one of its immediate predecessor and thus change $X^*$ to

$$(x_0 = 0, x_1 = 2, \ldots, x_k = 1, x_{k+1} = N - k - 1, x_{k+2} = 0, \ldots, x_{N-1} = 0).$$

As a result, the average does not change but variance will be reduced by $\frac{2k-1}{N}$. By induction on cycle size, the maximum reduction in variance will happen when the cycle has size of $k + 1$; leading to an $M^k A$ digraph. That is, we add an edge from a sink to one of its predecessors with a vertex accessibility that equals $k$. Then, $X^*$ will be changed to

$$(x_0 = 0, x_1 = 0, \ldots, x_k = k + 1, x_{k+1} = N - k - 1, x_{k+2} = 0, \ldots, x_{N-1} = 0).$$

The above solution represents an $M^k A$ digraph as shown in Figure 4.2. Accordingly, the variance will be reduced to $\left(1 - \frac{k+1}{N}\right) < \sigma_{acc}^D$ and the average accessibility would be $\mu_{acc}^* = (k+1)\left(1 - \frac{1}{N}\right)$ so that $[\mu_{acc}^*] = [\mu_{acc}^D] = k$. However, (1.4) requires that

$$\frac{N}{2} - 1 < k < \frac{3N}{2} - 1. \qquad (4)$$

It is obvious that (3) is much tighter than (4) since the upper bound of (4) can be waved because $\frac{3N}{2} - 1 > N$ and $k$ cannot be greater than $N - 1$ by definition. As a conclusion, $k$-accessible digraphs with $k \geq \frac{N}{2}$ must have a SCC of size $k + 1$ to have a minimal variance of vertex accessibility.

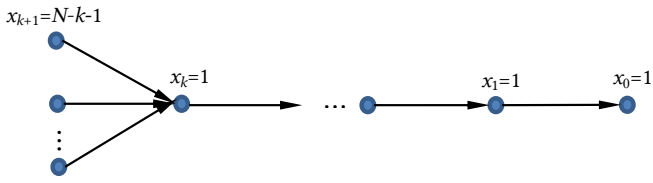
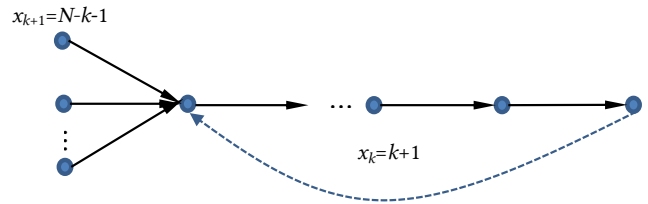

4.1. Spanning tree (Model 1)      4.2. $M^k A$

Figure 4. Optimal $k$-accessible digraphs

**Corollary 2:** Since a cycle is actually an SCC with minimal edge, the length of cycle in $M^k A$ digraph, i.e., $k + 1$, is an upper bound on the size of SCC in class of all $k$-accessible digraphs.

**Theorem 3**: Each M$^k$A digraph has a unique Giant SCC as core with order $k+1$ and without an outward edge.

**Proof**: Suppose the digraph contains more than one SCC of various sizes. Note that the size of each SCC cannot exceed $k$+1; otherwise, it is in contradiction with the *k*-accessible feature. Let $G'$ be the condensation digraph produced by contracting the SCCs into individual vertices. Now we argue that each SCC is a sink (without outward edge) in $G'$ of order $k$+1 to keep $\sigma_{acc}$ minimized.

Consider a candidate SCC containing $s$ vertices with one outward edge to vertex *i* as sink. Then, the partial variance of vertex accessibility related to the candidate SCC is given by $\sigma_{acc}^{SCC} = Q + s \times (k-s)^2 + k^2$ where $Q$ is the portion related to the inward edges to the SCC. Moreover, since $i$ is sink, we have $acc(i) = 0$, and $acc(v) = s \ \forall v \in$ SCC. Now we swap the SCC and vertex *i* so that SCC turns to the sink with an inward edge from vertex *i*. Thus, the revised partial variance is given by $\sigma_{acc}^{SCC'} = Q + s \times (k-s+1)^2 + (k-s)^2$ where $acc(i) = s$, and $acc(v) = s-1 \ \forall s \in$ SCC. Consequently, we have

$$\sigma_{acc}^{SCC'} = \sigma_{acc}^{SCC} + g(k,s),$$

where $g(k,s) = 1 + 2k - 2s(1+k) + s^2$. It is straightforward to show that parabola $g(k,s)$ has an optimal minimum value that equals $-k^2$ at $s = k+1$, resulting in $\sigma_{acc}^{SCC'} < \sigma_{acc}^{SCC}$. Thus, the lowest variance will be obtained when SCC is a sink of order $k+1$ in $G'$.

Finally, we show that the SCC is unique. Suppose not and thus there are some vertices in $G'$ having access to more than one SCC as sink in $G'$; otherwise, $G'$ is disjointed and contradicts with the original assumption. Thus, the abovementioned vertices have an accessibility equal or greater than $2(k+1)$. For instance, consider a digraph with two disjoint SCCs each of which has $k+1$ vertices and both have one common inward vertex. Thus, the order of digraph is $N = 2(k+1) + 1$. Thus,

$$\mu_{acc} = \frac{2(k+1)+(N-1)k}{N} = \frac{2(k+1)}{2(k+1)+1}(k+1).$$

Accordingly, $\mu_{acc} \to k+1$ when $k \to \infty$; contradiction with $G'$ being $k$-accessible. Thus, the uniqueness of SCC is a necessary and sufficient condition for $G'$ to be $k$-accessible. Some examples of $k$-accessible, as optimal solutions to Model (1), are shown in Figure 5. The number displayed close to each vertex represents $acc(v)$. The red circles represent the vertices with only output flow which are interpreted as correspondent sender (or receiver) in the CB network. As is evident in this figure, all digraphs have only one GSCC of size $k$+1.

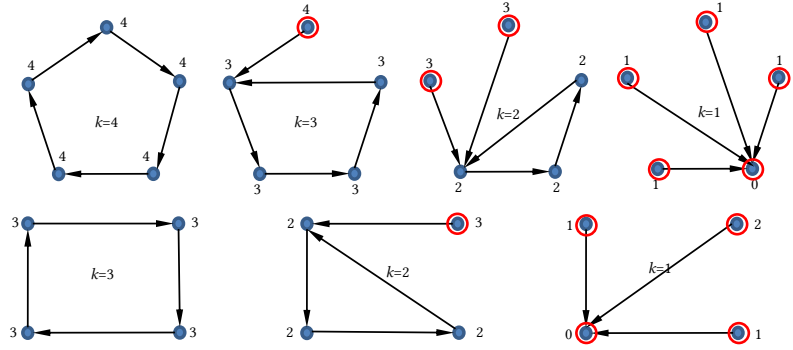

**Figure 5.** Examples of *k*-accessible digraphs with minimal $\sigma_{acc}$ – optimal solutions of Model (1)

**Corollary 3:** The accessibility variance in M$^k$A digraph, i.e., $\left(1 - \frac{k+1}{N}\right)$, is an upper bound on accessibility variance in all $k$-accessible digraphs.

Corollary 2 can be also re-verified through theorem 2. Suppose Corollary 2 is not true and thus GSCC of a $k$-accessible digraph contains more than $k+1$ vertices. Then, it is straightforward to show that $[\mu_{acc}] > k$, a contradiction with the original assumption.

Corollaries 2 and 3 provide a link between the above theorems and the real world. In practice, all sender correspondents do not necessarily have a tail in SCC, meaning that SCC is not the bridge set of the network. Moreover, all receiver correspondents are not necessarily belonging to SCC, meaning that SCC is not necessarily a sink without an outward edge. Thus, the variance of accessibility is not minimized in the real world. In fact, the above theorems and associated inferences shows that $\sigma_{acc}^{SCC}$ is a unique metric representing the magnitude of SCC and thereby the macro-structure of CB network, as depicted in Figure 6. Suppose the BFS procedure reveals that the given CB network of order $N$ is $k$-accessible. Thus, when $\sigma_{acc}^{SCC}$ approaches zero, all components of set $\Psi$ in Figure 6 will merge to a *Giant* SCC (GSCC) of order $k+1$; representing a sink without an outward edge to other banks. In contrary, when $\sigma_{acc}^{SCC}$ increases, the SCC turns into a bridge set of decreasing order $k' < k+1$. In this scenario, $\sigma_{acc}^{G} < 1 - \frac{k+1}{N}$ where $G$ is the core of network containing $SCC$ (see Figure 6). In this case, if the number of corresponding senders equals $k + 1 - k'$ (there is a path from all corresponding senders to SCC), SCC is certainly a bridge set.

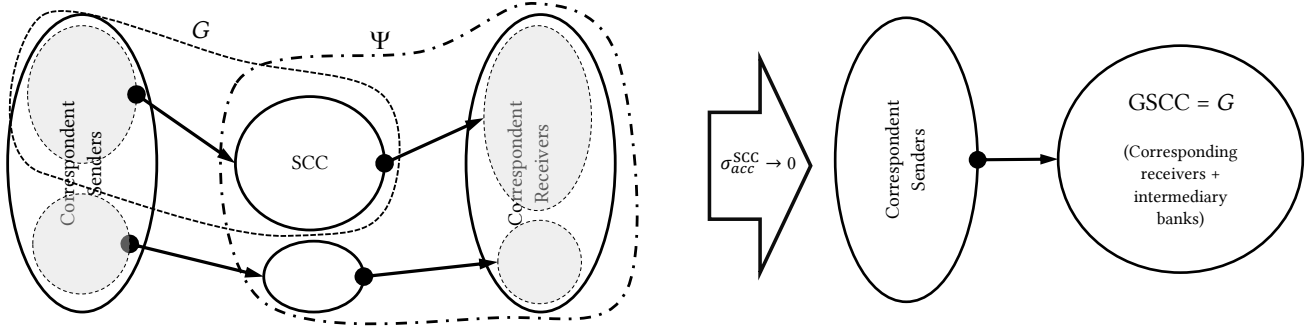

**Figure 6.** Macro structure of CB network as a function in terms of $\sigma_{acc}^{SCC}$

The best strategy for network size optimization is to look for the entities in SCC, the removal of which still preserves the strong connectivity of SCC. This strategy has two advantages: 1- reduction in computational efforts by limiting the exploration space to the SCC, 2- least negative effects on customers (only the direct customers of the removed entities will be affected). However, the diameter of SCC will be a constraint on optimization. Meaning that, by removing some elements of SCC, its diameter should not be increased; in other words, the CB network must remain a 6-Clan digraph.

### 3.2. Inferences on M$^p$C Digraphs

Through the following propositions and theorems, we study the situations under which the SCC of a $k$-accessible digraph could be $p$-Clan, given $p$ where $p > 1$. Recalling the definition of M$^p$C, an infinite diameter means the component is no longer an SCC. First, we find the minimal requirements a digraph must have to be $p$-Clan. To this end, an integer programming (IP) model is proposed to construct M$^p$C digraph and then the properties and topology of the digraph are studied through the optimality conditions. Then, we employ a theorem by Knyazev (1987) to find a tight upper bound on diameter of the digraph. This helps us to estimate the diameter without the need for deep exploration of the network structure by time-consuming traversal methods.

**Theorem 5:** The number of edges in minimal $p$-Clan digraph of order $N$ is

$$m^*(N,p) = \begin{cases} \left(1+\frac{2}{p}\right)(N-1) & \mod(2(N-1),p) = 0 \\ 2N - p & \mod(2(N-1),p) \neq 0 \end{cases}. \quad (5)$$

**Proof:** Let $D(V,E)$ be an M$^p$C digraph with $|V| = N > 3$ vertices and $m$ edges. It is obvious that for $p = 1$, the minimum number of edges to have a $p$-Clan digraph equals the number of edges in the associated complete digraph, i.e., $m^* = N(N-1)$. We propose the following Lemma to facilitate the calculation of $m^*$ when $p > 1$.

**Lemma 1:** The problem of minimizing edges is equivalent to the problem of maximizing the distances (shortest path) of length $p$ in a $p$-Clan digraph.

**Proof:** To proof Lemma 1, we propose an integer programming (IP) model to construct the $M^pC$ digraph, given $p$. Let $x_{ij}^l = 1$; if the distance between pair vertices $i$ and $j$ equals $l$ where $1 \leq l \leq N-1$; otherwise, $x_{ij}^l = 0$. Thus, in a $p$-Clan digraph, we have $\sum_{l=1}^{p} x_{ij}^l = 1 \ \forall i \neq j$. The problem is to minimize the number of edges, i.e., $\sum_i \sum_j x_{ij}^1$. The proposed IP is as follows.

$$\min \sum_i \sum_j x_{ij}^1, \tag{6.1}$$

s.t:

$$\sum_k \sum_{r=1}^{l-1} x_{ik}^r x_{kj}^{l-r} \geq (l-1) \times x_{ij}^l \quad \forall i \neq j, 2 \leq l \leq p, \tag{6.2}$$

$$\sum_{l=1}^{p} x_{ij}^l = 1 \quad \forall i \neq j, \tag{6.3}$$

$$x_{ij}^l \in \{0,1\}; \ x_{ii}^l = 0 \ \forall i, l. \tag{6.4}$$

The objective of (6.1) is to minimize the number of edges. Constraint (6.2) imposes that $x_{ij}^l = 1$ means $j$ is reachable by $i$ through $l-1$ intermediary vertices. The strict inequality of (6.2) indicates that several paths of same length may exist between the pair of vertices $(i,j)$. Also, this constraint avoids sub-tours and thereby guarantees the digraph to be connected. Constraint (6.3) avoids multiple distance assignments between pairs of vertices and, at the same time, does not allow the digraph diameter to exceed $p$. Finally, constraint (6.4) enforces the integrality of decision variables and avoids loops in the digraph.

Considering inequality constraints (6.2), we use the Karush-Kuhn-Tucker (KKT) conditions to study the optimality conditions. To this end, we consider the LP relaxation of model (6) where $0 \leq x_{ij}^l \leq 1$ and, thereby, the penalty term $x_{ij}^l(1 - x_{ij}^l) = 0 \ \forall i,j,l$ is added to ensure the integrality of the decision variables. Thus, the objective function and constraints are continuously differentiable in the relaxed model. To examine the KKT conditions, we construct the Lagrangian function as follow:

$$L = \sum_i \sum_j x_{ij}^1 + \sum_i \sum_j \sum_{l=2}^{p} \lambda_{ij}^l \left((l-1) \times x_{ij}^l - \sum_k \sum_{r=1}^{l-1} x_{ik}^r x_{kj}^{l-r}\right) + \sum_i \sum_{j \neq i} \theta_{ij} \left(\sum_{l=1}^{p} x_{ij}^l - 1\right) + \beta \sum_i \sum_j \sum_l x_{ij}^l (1 - x_{ij}^l),$$

where $-\infty < \theta_{ij} < +\infty$, $\lambda_{ij}^l \geq 0$ are KKT multipliers and $\beta$ is an arbitrary large positive penalty coefficient. The KKT regularity conditions imposes that

$$\begin{cases} \frac{\partial L}{\partial x_{ij}^1} = 1 - \sum_q \sum_{s=2}^{p}(\lambda_{iq}^s x_{jq}^{s-1} + \lambda_{qj}^s x_{qi}^{s-1}) + \theta_{ij} + \beta(1 - 2x_{ij}^1) = 0; \; \forall i \neq j & (7.1) \\ \frac{\partial L}{\partial x_{ij}^l} = -\sum_q \sum_{s=l+1}^{p}(\lambda_{iq}^s x_{jq}^{s-l} + \lambda_{qj}^s x_{qi}^{s-l}) + \lambda_{ij}^l(l-1) + \theta_{ij} + \beta(1 - 2x_{ij}^l) = 0; \; \forall i \neq j, 1 < l < p & (7.2) \\ \frac{\partial L}{\partial x_{ij}^p} = \lambda_{ij}^p(p-1) + \theta_{ij} + \beta(1 - 2x_{ij}^p) = 0; \; \forall i \neq j & (7.3) \\ \lambda_{ij}^l\left((l-1) \times x_{ij}^l - \sum_k \sum_{r=1}^{l-1} x_{ik}^r x_{kj}^{l-r}\right) = 0 \; \forall i \neq j, 1 < l \leq p & (7.4) \end{cases}$$

To lighten notations, we define $Q_{ij}^l = \sum_q \sum_{s=l+1}^{p}(\lambda_{iq}^s x_{jq}^{s-l} + \lambda_{qj}^s x_{qi}^{s-l})$ $\forall i \neq j, 1 < l < p$ where $Q_{ij}^l \geq 0$ and $Q_{ij}^p = 0$. Then, we draw the following inference from (7)

$$\begin{cases} x_{ij}^1 = \frac{1}{2}\left(1 + \frac{1}{\beta}(\theta_{ij} - Q_{ij}^1 + 1)\right) \\ x_{ij}^l = \frac{1}{2}\left(1 + \frac{1}{\beta}(\lambda_{ij}^l(l-1) + \theta_{ij} - Q_{ij}^l)\right); \; 1 < l < p. \\ x_{ij}^p = \frac{1}{2}\left(1 + \frac{1}{\beta}(\lambda_{ij}^p(p-1) + \theta_{ij})\right) \end{cases} \quad (8)$$

Thus, under the optimal conditions, the objective function (6.1) can be rewritten in terms of KKT multipliers as

$$\sum_i \sum_{j \neq i} x_{ij}^1 = \frac{1}{2}\left(N(N-1) + \frac{1}{\beta}\sum_i \sum_{j \neq i}(\theta_{ij} - Q_{ij}^1 + 1)\right).$$

The above term reveals that the objective function of dual problem is equivalent to maximizing $\sum_i \sum_{j \neq i}(\theta_{ij} - Q_{ij}^1)$. Suppose $\beta$ is large enough so that $x_{ij}^l(1 - x_{ij}^l) = 0$ $\forall i, j, l$ is satisfied. Then, the possible scenarios for the decision variables under the optimal conditions are

$$\begin{cases} x_{ij}^l = 1 \Leftrightarrow \lambda_{ij}^l(l-1) + \theta_{ij} - Q_{ij}^l = \beta \\ x_{ij}^l = 0 \Leftrightarrow \lambda_{ij}^l(l-1) + \theta_{ij} - Q_{ij}^l = -\beta \end{cases} \quad \forall l > 1.$$

If $x_{ij}^1 = 0$, then $Q_{ij}^1 - \theta_{ij} = \beta + 1$ meaning that $\theta_{ij} < Q_{ij}^1$. Moreover, Constraint (10.2) enforces $\sum_{l=1}^{p} x_{ij}^l = 1$ which means there exist $l^* > 1$ where $x_{ij}^{l^*} = 1$ and thereby (8) imposes $\lambda_{ij}^{l^*}(l^* - 1) + \theta_{ij} - Q_{ij}^{l^*} = \beta$. As a result, $x_{ij}^1 = 0$ enforces

$$\theta_{ij} = Q_{ij}^1 - \beta - 1 = \beta - \lambda_{ij}^{l^*}(l^* - 1) + Q_{ij}^{l^*}$$

$$\Rightarrow Q_{ij}^1 - Q_{ij}^{l^*} = 2\beta - \lambda_{ij}^{l^*}(l^* - 1) + 1.$$

Since $\beta$ is a large positive value, we have $Q_{ij}^1 > Q_{ij}^{l^*} \geq 0$. Knowing that $\theta_{ij} < Q_{ij}^1$, the partial term $(\theta_{ij} - Q_{ij}^1)$ in dual objective will be maximized if $\theta_{ij} \to Q_{ij}^1$ and $Q_{ij}^1 \to 0$ and consequently $Q_{ij}^{l^*} \to 0$. Thus, recalling $Q_{ij}^p = 0$, $l^* = p$ is always an optimal choice. As a conclusion, the optimality enforces either $(x_{ij}^1 = 1, \; x_{ij}^l = 0 \; \forall l > 1)$ or $(x_{ij}^1 = 0, \; x_{ij}^p = 1)$ and thereby

$$\min \sum_i \sum_j x_{ij}^1 \equiv \max \sum_i \sum_{j\neq i} x_{ij}^p. \square$$

For discussion on sufficient optimality of Conditions (7), refer to Appendix A.

Now we can prove Theorem 5 by induction on $p$ where, according to Lemma 1, the objective is to maximize the distances of length $p$. To do so, we consider the following scenarios for $p$ to execute the induction.

**Scenario 1**: Suppose $p = 2$, then we show that the star topology with a single hub and bi-directed edges has a minimum number of edges and maximum number of distances of length 2. First, we rewrite (6.2) for $p = 2$,

$$\sum_k x_{ik}^1 x_{kj}^1 \geq x_{ij}^2 \quad \forall i \neq j,$$

and thus,

$$\sum_{j\neq i} x_{ij}^2 \leq \sum_{j\neq i} \sum_k x_{ik}^1 x_{kj}^1 = \sum_k x_{ik}^1 \left(\sum_{j\neq i} x_{kj}^1\right) \quad \forall i.$$

We can state $\sum_{j\neq i} x_{kj}^1 = O_k - x_{ki}^1$; where $O_k$ is out-degree of vertex $k$. Since $D$ is 2-Clan, $\sum_{l=1}^2 \sum_{j\neq i} x_{ij}^l = O_i + \sum_{j\neq i} x_{ij}^2 = N - 1 \; \forall i$, must be ensured. Thus,

$$N - 1 = O_i + \sum_{j\neq i} x_{ij}^2 \leq O_i + \sum_k x_{ik}^1 (O_k - x_{ki}^1) = O_i + \sum_{k\neq i} x_{ik}^1 O_k - \sum_k x_{ik}^1 x_{ki}^1.$$

The term $\sum_k x_{ik}^1 x_{ki}^1$ represents the number of 2-cycle components that have a tail in vertex $i$. Let $y_{ij} \equiv x_{ij}^1$, then Model (6) is equivalent to the following MIP model

$\min \sum_i \sum_j y_{ij},$ \hfill (9.1)

Subject to

$O_i + \sum_{k\neq i} y_{ik} O_k - \sum_k y_{ik} y_{ki} \geq N - 1 \; \forall i,$ \hfill (9.2)

$O_i = \sum_{j\neq i} y_{ij} \; \forall i,$ \hfill (9.3)

$y_{ij} \in \{0,1\} \; \forall i.$

The optimality conditions enforce the strict equality for constraints (9.2) so that the only feasible solution is

$$y_{ih} = 1, \; y_{hi} = 1, y_{ij} = 0 \; \forall i, j \neq h.$$

The above solution indicates $O_i = \sum_k y_{ik} y_{ki} = 1 \; \forall i \neq h$ and $O_h = \sum_k y_{hk} y_{kh} = N - 1$; representing the star topology with vertex $h$ as hub.$\square$

**Scenario 2**: Suppose $p > 2$, let any arbitrary node to be hub and divided the rest $N - 1$ nodes into $L$ cycles (folds) each of which contains $K$ nodes so that $\mathrm{mod}(N - 1, \; K) = 0$. If each cycle is contracted to a single node and then connected to the hub through a bi-directed edge, the obtained structure would be a star topology same as

Scenario 1. Therefore, this structure is also optimal for $p = 2K$ with $L \times (K + 1) = \left(1 + \frac{2}{p}\right)(N - 1)$ edges where $\mathrm{mod}(2(N - 1), p) = 0$. In this case, the hub has a degree equal to $N - p + 1$ and the rest of nodes have a degree equal to one.

Suppose $\mathrm{mod}(2(N - 1), p) \neq 0$. Then, similar to the former case, we try to construct a semi-star topology where the hub has a degree equal $N - p + 1$ and the rest of the nodes have a degree equal one. To do so, a large cycle of size $p$ is created and the rest $N - p$ vertices are attached to a single vertex of the cycle through bi-directed edges. This structure has $p + 2(N - p) = 2N - p$ edges (each bi-directed edge represents a 2-cycle).

Theorem 5 can be also verified through the experimental results. To this end, we solve Model (6) for $N = 3, 4, \ldots, 10$ and $p = 1, 2, \ldots, N - 1$ using Branch-and-Bound method. The objective values are summarized in Table 1 and the solutions for $N = 5, 6$ are schematically shown in Figure 7. Since Model (6) is nonlinear due to quadratic constraint (6.2), the global optimum cannot be guaranteed. However, the solutions reported in Table 1 are either exactly the same as what we claimed in Theorem 5 or extremely close. Moreover, the resulted 2-Clan digraphs have a single-hub star topology as claimed in Lemma 1 and schematically shown in Figure 7.

Table 1. Model 10 – z value

| ↓N  p→ | 1  | 2  | 3  | 4  | 5  | 6  | 7  | 8  | 9  |
|---|---|---|---|---|---|---|---|---|---|
| 3  | 6  | 3  |    |    |    |    |    |    |    |
| 4  | 12 | 6  | 4  |    |    |    |    |    |    |
| 5  | 20 | 8  | 7  | 5  |    |    |    |    |    |
| 6  | 30 | 10 | 9  | 8  | 6  |    |    |    |    |
| 7  | 42 | 12 | 11 | 9  | 9  | 7  |    |    |    |
| 8  | 56 | 14 | 13 | 11 | 10 | 10 | 8  |    |    |
| 9  | 72 | 16 | 16 | 12 | 12 | 11 | 11 | 9  |    |
| 10 | 90 | 18 | 18 | 14 | 13 | 13 | 12 | 12 | 10 |

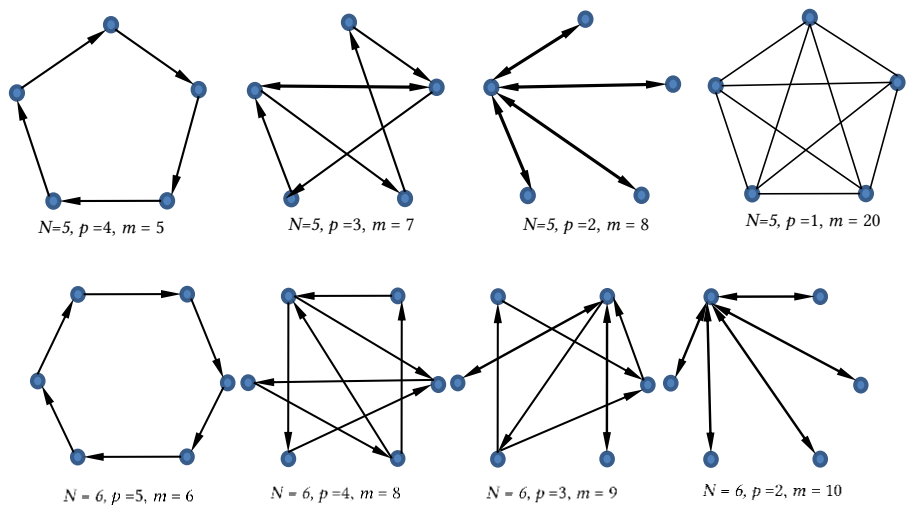

Figure 7. Examples of p-Clan digraphs with minimal edges – optimal solutions of Model (6)

**Corollary 4:** If a bank branch belongs to SCC, then its parent bank also does.

**Proof:** It is straightforward since there is a 2-cycle relationship between the parent bank and all of its branches.

**Knyazev's Theorem [21]:** Let $D(V,E)$ be a strongly connected *Eulerian* digraph (link symmetry - $O_i = I_i$ $\forall i \in V$) with $N$ vertices and minimum out-degree $\delta$, $2 \leq \delta \leq \frac{N}{2}$. If $D$ contains no 2-Cycle then, the diameter of $D$ is at most $\frac{5}{2\delta+2}N$. Moreover, if $D$ is $\delta$-regular digraph ($O_i = I_i = \delta$ $\forall i \in V$), then the diameter of $D$ is greater than $\frac{4}{2\delta+1}N - 4 + \frac{1}{2\delta+1}$.

Dankelmann (2005) showed that Knyazev's upper bound can be improved to $\frac{4}{2\delta+1}N - 4$ [22]. Knyazev's theorem requires $D$ to be *regular* (without 2-cycle), Eulerian and $2 \leq \delta \leq \frac{N}{2}$. Thus, to implement Knyazev's Theorem for our study, we propose the following propositions to meet these requirements.

**Proposition 1:** Any strongly connected digraph $D(V,E)$ can be reduced to a *regular* strongly connected digraph by contracting each 2-cycle component into a single vertex.

**Proposition 2:** Any strongly connected digraph $D(V,E)$ is extended to a strongly connected *Eulerian* digraph using the following procedure:

Let $d_i = \max\{I_i, O_i\}$ be the degree of vertex $i \in V$ in the extended Eulerian digraph. To ensure the equality of in-degree and out-degree for each vertex $i \in V$, we insert the set of nominal vertices $\Omega$ where $|\Omega| = \max_{i \in V}\{|I_i - O_i|\}$. Then, $d_i - O_i$ outward edges are added from vertex $i$ to some vertices in $\Omega$; if $d_i = I_i$. Likewise, $d_i - I_i$ inward edges are added from some vertices in $\Omega$ to vertex $i$; if $d_i = O_i$. Thus, the total inward edges from $V$ to $\Omega$ is $\sum_{i \in V}(d_i - O_i)$ and the total outward edges from $\Omega$ to $V$ is $\sum_{i \in V}(d_i - I_i)$. It is well-understood that $\sum_{i \in V} I_i = \sum_{i \in V} O_i = m$ and therefore $\Omega$ has the same in-degree and out-degree equals $\sum_{i \in V} d_i - m$. Since $D$ is strongly connected, it is straightforward that $\sum_{i \in V} d_i - m = 2|\Omega|$; then, the equality of in-degree and out-degree for each vertex in $\Omega$ is also guaranteed.

**Proposition 3:** The minimum out-degree in Eulerian digraph generated by Proposition 2 is maximized using the following procedure:

Let $\varsigma_i \subset \Omega$ be the set of all vertices in $\Omega$ which are not adjacent with $i \in V$; after applying Proposition 2.

DO

    Set $\delta = \min_i\{d_i\}$ and $j = \arg\min_i\{d_i\}$.

    Find a cooperative vertex $q \in V$ where $|\varsigma_j \cap \varsigma_q| \geq 2$ and $d_i \leq \frac{N}{2}$. If such a cooperative vertex cannot be found then STOP; else

        o  Select vertices $v, w \in \varsigma_j \cap \varsigma_q$ and add edges $i \to v$, $w \to i$, $j \to w$, $v \to j$ to ensure all vertices $i, j, v, w$ have the same incremental in- and out-degree.

        o  Set $d_j \leftarrow d_j + 1$, $d_q \leftarrow d_q + 1$, $\varsigma_j \leftarrow \varsigma_j/\{v,w\}$, $\varsigma_q \leftarrow \varsigma_q/\{v,w\}$.

LOOP

The above procedure has a linear complexity. The artificial increase in the number of vertices in Proposition 2 negatively affects the Knyazev's upper bound. However, Proposition 3 enables us to compensate this effect by increasing the minimum out-degree of the digraph. The above propositions enable us to set an explicit upper bound on diameter of the CB network; only requiring the out-degree of the entities.

### 3.3. Summary of Inferences

The findings of Section 3 can be summarized as follows; providing some insights for designing and optimizing the CB networks.

o    Possible scenarios for CB network optimization:

- If the termination of relationship is allowed (as is supposed in this study)

    o  and the suspicious entities are known in advance, we encounter a what-if analysis; how the network's characteristics are affected if some specific entities are dropped off.

    o  If the objective is to improve the overall risk profile relative to the returns of the network, then certain benefits may be realized by minimizing the SCC size – finding a maximal subset of entities belonging to SCC, the removal of which still preserves the strong connectivity without an increase in diameter (see Figure 8 for examples). Solving this problem is beyond the scope of this study and recommended as future work. As a guideline, by associating a cost coefficient per entity, the problem can be formulated as a class of Asymmetric Salesman Problem with some side constraints to preserve the SCC's characteristics.

- If the relationship should not be terminated but can be restricted, the problem is equivalent to finding an SCSS (*Strongly Connected Spanning Subgraph*) inside the SCC without increasing in diameter. This is equivalent to finding M$^p$C digraph associated with SCC if $p$ is given. Theorem 5 provides an explicit lower bound for this problem and Model (6) can be modified to solve this problem. Khuller et. al (1995) showed that when the circumference is three, the classical SCSS problem is as hard as Bipartite Matching. When it is five, the problem is NP-hard. When it is seventeen, the problem is AX-SNP-hard [17]. In Appendix B, we provide some insights to estimate the circumference of $p$-clan digraphs; however, we do not discuss the complexity.

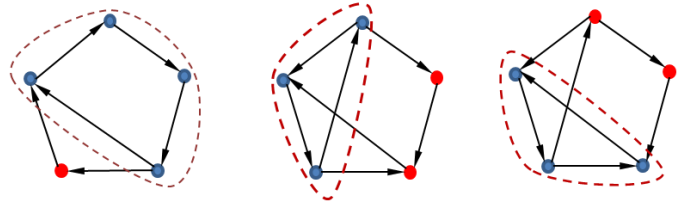

**Figure 8.** The red vertices represent the subset, the removal of which leaves the rest of the component strongly connected without increasing its diameter

- CB network $D(V,E)$ is $k$-acccessible ($k \geq \frac{N}{2}$), if it contains core $G \subseteq V$ where $(v) \approx k \; \forall v \in G$. If so,

  - There is a unique $SCC(V_0, E_0)$ with $|V_0| = n$ vertices and $|E_0| = m$ edges where $|V_0| \subseteq G$ and $acc(v) = k \; \forall v \in |V_0|$. The order of SCC is at most $k + 1$ ($n \leq k + 1$) (Corollary 2).

  - If $G = V$ then the CB network itself is a GSCC

  - If $\sigma_{acc} \leq 1 - \frac{k+1}{N}$ for all $v \in G$, the SCC is the bridge set of the network (Corollary 3).

  - Since $k \geq \frac{N}{2}$, any entity $v \in V$ where $acc(v) > k$ has outward edge(s) to $SCC$. These vertices connect (directly or indirectly) the senders to the SCC (*sender correspondents*) and therefore cannot be eliminated from the network.

  - Likewise, any entity $v$ where $acc(v) < k$ connects the rest of the network to receivers (*receiver correspondents*) and thus also cannot be eliminated from the network.

  - Given value $p$ in Theorem 5,
    - If $m < m^*(N,p)$, SCC cannot be $p$-Clan.
    - If $m \approx m^*(N,p)$ and SCC has semi-star or multi-fold topology then it should be $p$-Clan.

- If $m > m^*(N,p)$, let $SCC'(V_0', E_0')$ be the Eulerian subgraph associated with $SCC$ after applying Propositions 1,2, and 3. Then, the diameter of $SCC$ is at most.

$$\frac{4}{2\delta+1} n' - 4,$$

where $n' = |V_0'| + |\Omega|$, $\delta = \min_{i \in V'}\{d_i\}$ and $|\Omega| = \max_{i \in V'}\{|I_i - O_i|\}$.

## 4. EXPERIMENTAL RESULTS

This section is dedicated to verification of the theorems in Section 3 using samples of Scotiabank SWIFT payments records, a dataset of 8.7 million cross-border payments over the 2018 fiscal year. The SWIFT messages allow us to track the flow of individual payments from sender to receiver passing through the different transactional banks. The *original* CB network extracted from this dataset consists of about 12,000 branches belonging to about 2,000 various financial institutions, serving nearly 200,000 customers across the world.

*Network Sampling*

For the purpose of benchmarking, we take a few smaller samples in the form of subnetworks from the original CB network. Each sample represents a pool of payments that went through all branches of a candidate bank distributed across the world. This strategy results in high diversity of *assortativity* and thereby in less bias in sampling. The assortativity measures the similarity of connections in the graph with respect to the node degree. This sampling strategy is not biased toward high-degree nodes, which is an issue with many graph traversal techniques (e.g., BFS, DFS, Forest Fire, Snowball Sampling, RDS) [23]. Thus, the subnetworks created using the above strategy are plausible standalone scale-free networks because the degree distribution follows a power law, as depicted in Figure 9.

The average order of bank subgraph in sampled subnetworks is 3000 entities. The average number of edges is 1.5 times that of the entities, representing a sort of sparse subnetwork. Since the theorems proved in Section 3 are based on minimal digraphs, the sparse (rather than dense) samples are preferred for verification purposes. The ratio of the number of intermediary banks to the number of correspondent banks is less than 5%. The assortativity of digraph in all samples is negative; representing absolute dissimilarity between degrees of adjacent vertices.

*Sample 1*: The first sample is a subnetwork with 2212 vertices and 3878 edges. The histograms of inward/outward degree of vertices are displayed in Figure 9. While the distribution of inward and outward degrees is highly right-skewed (power law), their range is also relatively high. This is due to our sampling policy in which all transactions went through a few major branches of a candidate bank. The vertices with zero inward degree represent sender correspondents and those with zero outward degree represent receiver correspondents.

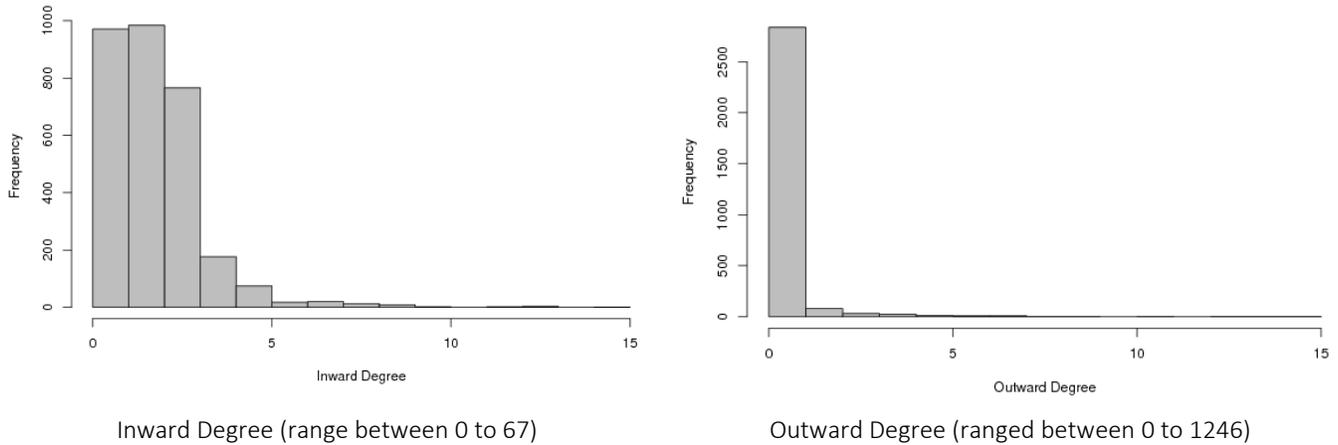

Inward Degree (range between 0 to 67)     Outward Degree (ranged between 0 to 1246)

**Figure 9**. Inward/outward degrees of vertices – Sample 1

The histogram of vertex accessibility calculated using the BFS procedure is shown in Figure 10. The accumulated mass over extreme values indicates a $k$-accessible subnetwork with $k = 1840$. That is, there is a core of 100 entities having access to, on average, 1840 other entities where $\sigma_{acc} = 4.75$. Since $\sigma_{acc} > 1 - \frac{k+1}{N} = 0.1681$, according to Corollaries 2 and 3, the SCC is neither a bridge set nor a sink in the network. 26 out of 100 entities construct SCC with accessibility equal to 1838 and 50 out of 100 entities have a tail in SCC with an equal accessibility of 1839. The existence and uniqueness of the SCC is verified using *components(.)* function in the *igraph* package in R. This function is used to detect all possible strongly connected components in the network using *Tarjan*'s algorithm based on deep-first search (DFS) method with a linear complexity. There are 23 entities that have accessibility greater than $k$ ranging from 1840 to 1874. Moreover, there are 106 entities with accessibility ranging between 1 and 26, meaning that none of them have a tail in SCC. The rest of the entities have zero accessibility, representing receiver correspondents.

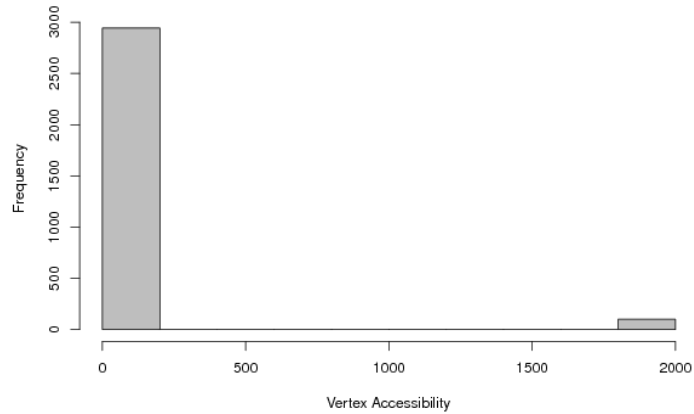
**Figure 10.** Vertex accessibility– Sample 1

The assortativity of the subnetwork is -0.214; calculated using the *igraph* package in R. As pointed out earlier, a negative assortativity represents very good diversity among the degree of vertices. The reason for this promising value is our sampling strategy where we choose the candidate banks whose branches are well-distributed geographically. The entities in Sample 1 are geographically distributed in 86 countries and 5 continents.

The above analysis reveal that the search space for network optimization could be reduced from 3043 entities and 3878 edges (subnetwork shown in Figure 12) to only 27 potential entities and 64 edges (SCC shown in Figure 13). As discussed at the end of the last section, we are looking for the entities belonging to SCC, the removal of which preserves the strong connectivity and SCC's diameter. If such entities exist, the parent bank and thereby all associated branches can be removed from the network (Corollary 4) without negative effects on the customers.

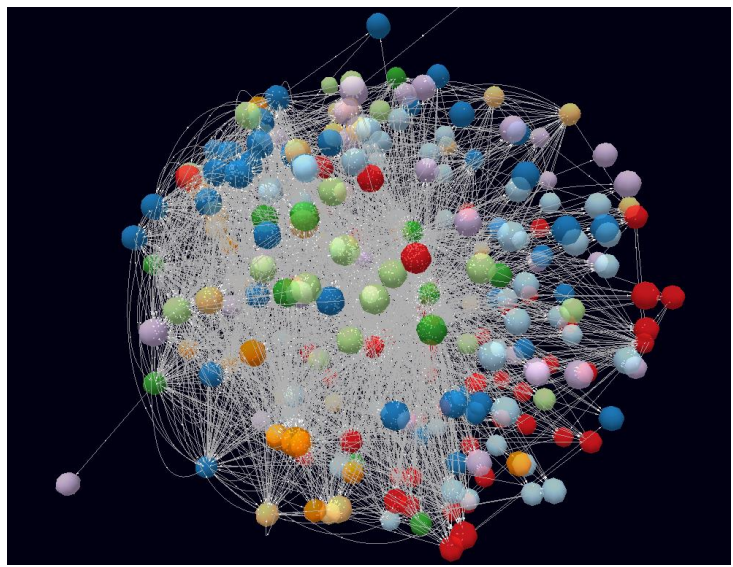
**Figure 11.** Geographical distribution of the original CB network (each node represents a country, the color represents the continent, and the links represent the money flow among different countries)

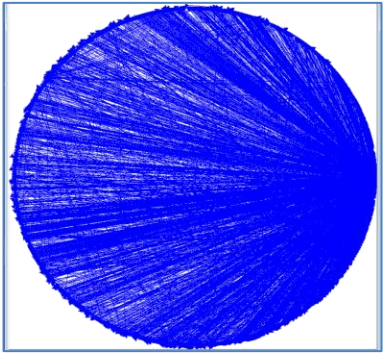 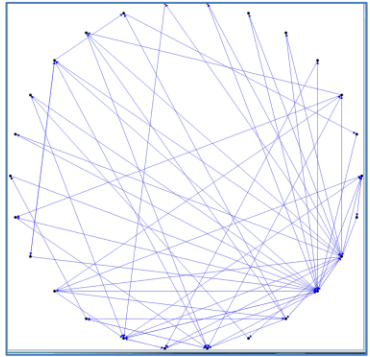 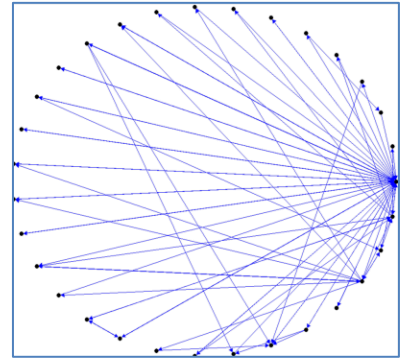

**Figure 12**. Subnetwork- Sample 1    **Figure 13**. GSCC – Sample 1    **Figure 14**. GSCC –Sample 2

Using *diameter*() function in the *igraph* package, we find the diameter of the entire subnetwork is 7, while the diameter of the CSS is 6, indicating a 6-Clan digraph.

To examine Knyazev's upper bound, first we regularize SCC using Proposition 1. This results in a reduced subgraph of order 23 due to the elimination of four 2-cycles. Then, we turn the subgraph into an Eulerian subgraph using Propositions 2 and 3. To this end, we add $max_i\{|I_i - O_i|\} = 10$ nominal entities and necessary edges to the subgraph to ensure the equality of in-degree and out-degree for all entities. The outcome is an Eulerian subgraph of order 23+10 = 33 and minimum out-degree $\delta = 6$. Thus, Knyazev's upper bound is $\frac{4}{2\times 6+1}(33) - 4 = 6.15$ which is nearly same as the diameter calculated by the *igraph* package.

Moreover, according to Theorem 5, the SCC must have at least $m^*(23,6) = 60$ edges to be eligible for being 6-Clan, while currently it has 64 edges. Thus, the optimal entities for reduction are those the removal of which does not reduce the number of edges by more than 64-60 = 4.

*Sample 2*: The second sample is a subnetwork of order 3,000 with 3316 edges and assortativity coefficeint -0.447. This subnetwork is a 2850-accessible network with a core of order 138 entities have access to, on average, 2850 other entities where $\sigma_{acc} = 1.390$. 31 out of 138 entities construct the GSCC shown in Figure 14. The inward degree, outward degree and accessibility of vertices follow similar patterns shown in Figures 9 and 10. The GSCC has diameter equals 5; however, Knyazev's upper bound is $\frac{4}{2\times 6+1}(29+5) - 4 = 6.2$. Theorem 5 imposes at least $m^*(31,5) = 57$ edges for GSCC to be 5-Clan and currently it has 75 edges. Thus, the optimal entities for reduction are those the removal of which does not reduce the number of edges by more than 18.

## Structural Analysis of Original CB Network

The original CB network has 33,901 nodes with 192,855 edges and assortativity coefficeint -0.318. The network is 26,487-accessible with a core of order 23,767 and a GSCC of order 16,406 where $\sigma_{acc} = 7.829$ which is significantly greater than the threshold $1 - \frac{k+1}{N} = 0.218$. Then, we can say GSCC play the role of bridge set in the network due to a large $\sigma_{acc}$ compared to the threshold. Since $k \geq \frac{N}{2}$, any entity with accessibility greater than $k$ has outward edge(s) to GSCC. That is, 23,767 - 16,406 = 7,361 entities represent the corresponding senders and the rest (33,901 – 23,767 = 10,134) represent the corresponding receivers. As depicted in Figure 15a, there is a significant positive correlation between inward and outward degree of entities in GSCC whereas each one follows a power law distribution, as shown in Figure 9. Moreover, the indgree-to-outdegree ratio (Figure 15b) indicates a sort of link symmetry (indegree ≈ outdegree) in GSCC. However, the non-GSCC entities (more than half of the network - Figure 15b) are corresponding senders and receivers without spesific in/out-degree pattern. The link Symmetry increases the overall connectivity of the network and reduces its diameter. As an insight, the GSCC in CB network behave like a sotial network (scale-free) rather than web (hub-and-spoke structure).

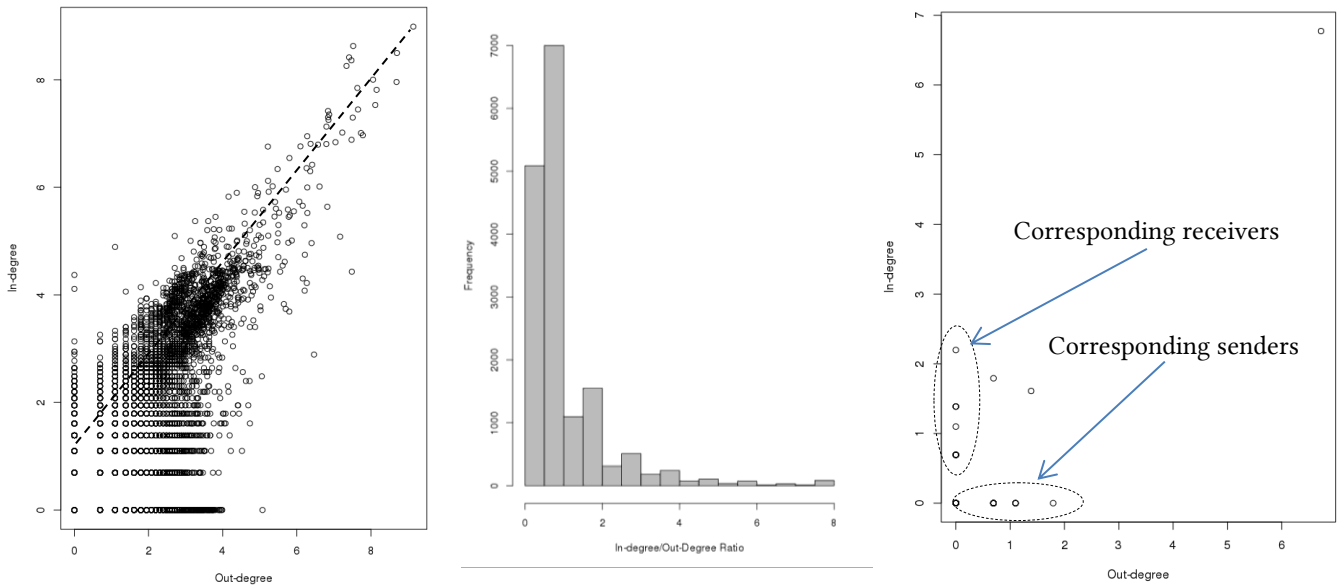

| 15a. Indegree vs. outdegree (log-scaled) GSCC | 15b. Indegree-to-outdegree ratio GSCC | 15c. Indegree vs. outdegree (log-scaled) non-GSCC |

**Figure 15.** Inward vs. outward degree of entities in original CB network

The diameter of the entire network is 7 and the diameter of the GCSS is 6. The GSCC must have at least $m^*(16406, 6) = 21,705$ edges to be eligible for being 6-Clan; while, it currently has 166,791 edges. So, there is enough room for the optimization; without affecting the GSCC's diameter.

## 5. CONCLUSIONS

This research work is the first to provide a theory-based view of CB network and provides theoretical foundation to explain the network optimization scenarios. The provided theorems and related inferences enable us to understand the structural properties of CB networks and reduce the search space for network optimization; avoiding the need for deep exploration of the network. We discovered a remarkable feature of CB networks, $k$-accessibility, where the network contains a maximal set of entities, i.e., core, that have access to, on average, $k$ other banks with a negligible variance. The key reason for this behaviour is the child-parent relationship between branches and their head offices. This feature has been verified through the network sampling. We showed that any $k$-accessible network has a Giant Strongly Connected Component of order $k+1$ and thus our search for the potential evaluation and remediation should be restricted to this component instead of the entire network. Additionally, we argued that the diameter of the network must be a concern through the optimization phase to ensure a reasonable length for alternative transaction routes after removing some entities.

We employed the mathematical programming approach to prove several theorems through studying the optimality conditions of the proposed models. We showed that constructing the minimal $k$-accessible and $p$-Clan digraphs can be formulated as MIP and quadratic IP models respectively.

We investigated the structural properties of CB networks using a binary (unweighted) adjacency matrix. However, in future work, a real-valued adjacency matrix may represent the amount or volume of historical transactions among entities, indicating the connection's strengths. Having this, we can measure the importance of each entity in the network through assigning relative centrality/prestige score (e.g., betweenness, closeness, eigenvector,…) to each entity. The entities with the lowest score can be considered for evaluation or remediation. Alternatively, one may employ an unsupervised clustering approach to classify the entities in terms of a set of features obtained from the information existing in an adjacency matrix, properties extracted from the network structure, and various centrality scores. Moreover, more investigation on the structural properties of CB networks such as studying redundant edges, cycles and Pancyclicity may help us to detect some suspicious structures that may represent anomalous activities. The inferences discussed in this study are also applicable in directed-graph applications beyond the particular problem of financial networks.

## Appendix A: Sufficient Optimality Conditions

The system of equations (7) contains the first-order necessary conditions for a solution to be optimal. These conditions will also represent the sufficient optimality conditions, if $(-z)$ is a concave function, inequality constraints (6.2) are convex functions and equality constraints (6.3) are affine functions. Even though $z$ is a linear function (and thereby concave), (6.2) is not a convex function and therefore we have to verify the *Slater* condition. In convex optimization problems with differentiable objective and constraints, the *Slater* condition imposes that if there is a feasible solution so that Constraints (6.2) hold with strict inequality (feasibility), then KKT conditions provide both necessary and sufficient conditions for optimality. Slater's condition guarantees the strong duality and implies that the optimal duality gap is zero and the dual optimum is attained. As pointed out earlier, constraints (6.2) hold strict inequality when the path between the pair of vertices is not unique. This is a case in complete digraphs (1-Clan) where there are multiple paths of same length between pair vertices. The complete digraph is always a feasible solution to Model (6) for any $p$. Hence, Slater's condition is always held by Model (6). As a result, the system of equations (7) represents both necessary and sufficient optimality conditions.

## Appendix B: Estimation on circumference of $M^pC$ digraphs

**Theorem:** The circumference of minimal $p$-Clan digraph $(p > 2)$ is at most $p + 1$.

**Proof:** Scenario 2 of Theorem 5 reveals that the circumference is 2 for $p = 2$ (recall that each bi-directed edge represents a 2-cycle). Suppose $p > 2$, then according to the proposition 2.1 in Khuller et al., (1995): "*If the circumference of an strongly connected digraph of order $N$ is $l$, then any SCSS has at least $(N - 1)\frac{l}{l-1}$ edges*" [24]. Suppose circumference of $M^pC$ digraph is $p + 1$; thus, any SCSS extracted from the digraph has at least $m^{SCSS} = (1 + \frac{1}{p})(N - 1)$ edges which is an extremely tight lower bound on $m^*(N, p)$ from Theorem 5. Therefore, the circumference should be at most $p + 1$; otherwise $m^{SCSS} > m^*(N, p)$ which is a contradiction with the definition of SCSS.

**Corollary:** The circumference of any $p$-Clan digraph $(p > 2)$ is at most $p + 1$.

**Woodall's Theorem**: "*Given undirected graph $G$ with $N$ vertices, $m$ edges $(m \geq N)$ and circumference $c$, then*

$$m \leq \frac{1}{2}(N - 1)c - \frac{1}{2}r(c - r - 1),$$

*where $r = r(N - 1, c - 1)$. Given positive integers $a$ and $b$, $r(a, b) = a - b\lfloor a/b \rfloor$, the remainder on division $a$ by $b$*" [19].

**Proposition:** Woodall's Theorem can be improved for digraphs as $m \leq (N - 1)c - r(c - r - 1)$. Thus, a lower bound on circumference is obtained as

$$c \geq \frac{m - r(r+1)}{N - (r+1)}.$$

Suppose the digraph is $p$-Clan. Since $r$ is a positive integer number and according to Theorem 5 we have $m \geq m^*(N,p) > N$, the left hand side of above term will be maximized whenever $r = 0$. Thus, the lower bound on the circumference of any digraph can be improved to $c \geq \frac{m}{N-1}$. As a conclusion, the circumference of any $p$-Clan digraph is bounded as

$$\frac{m}{N-1} \leq c \leq p + 1.$$